# An Immediate Update Strategy of Multi-State Constraint Kalman Filter

Qingchao Zhang, Wei Ouyang, *Member, IEEE*, Jiale Han, Qi Cai, Maoran Zhu, *Member, IEEE* and Yuanxin Wu, *Senior Member, IEEE*

*Abstract*— The lightweight Multi-state Constraint Kalman Filter (MSCKF) has been well-known for its high efficiency, in which the delayed update has been usually adopted since its proposal. This work investigates the immediate update strategy of MSCKF based on timely reconstructed 3D feature points and measurement constraints. The differences between the delayed update and the immediate update are theoretically analyzed in detail. It is found that the immediate update helps construct more observation constraints and employ more filtering updates than the delayed update, which improves the linearization point of the measurement model and therefore enhances the estimation accuracy. Numerical simulations and experiments show that the immediate update strategy significantly enhances MSCKF even with a small amount of feature observations.

## I. INTRODUCTION

Recent decades have witnessed the thriving researches and applications of the visual-inertial odometry (VIO) in numerous integrated navigation systems, such as the unmanned arial vehicles [1][2], autonomous underwater/subterrain vehicles [3][4], autonomous driving [5], etc. The VIO algorithms are typically classified as the filtering-based and optimization-based algorithms [6], and as shown in the benchmark [7] the optimizers perform much better than the filters in terms of accuracy and robustness at the sacrifice of more computational costs. In computationally constrained platforms, the multi-state constraint Kalman filter (MSCKF) [8] has been known for its high efficiency [9] and its Kalman filtering realization is friendly to practitioners. Nevertheless, due to the lack of iterations on the linearization points, the original MSCKF was confronted with the inconsistency problem, and its estimation performance is unsatisfactory especially in long-duration missions [10]-[12].

Since the proposal of the well-known MSCKF, tremendous efforts have been invested to ameliorate the filtering consistency either by maintaining the intrinsic observability property of the VIO system or by mitigating linearization errors of the Kalman filter. Sun [15] adopted the stereo camera in S-MSCKF and showed greater robustness and accuracy improvement over the monocular VIO. The robo-centric formulation and iterated filtering were proposed in ROVIO [16] to improve the accuracy of linearization points. Huang [12] proposed the observability-constrained extended Kalman filter for tackling the spurious observability in MSCKF caused by the linearization errors, and the usage of first-estimate Jacobians [13][14] served the similar purpose. In recent years, the on-manifold Kalman filtering techniques have been applied to deal with the linearization errors in MSCKF, showing significant improvements on the estimation consistency and accuracy [17][18]. For instance, Brossard [19] and Wu [20] applied the Invariant Kalman filtering to MSCKF, which models the system states on the matrix Lie group and the resultant indirect Kalman filter naturally obeys the observability constraints due to the independence of estimated states. The equivariant filter-based algorithm MSCEqF [21] also achieved highly consistent estimation even when the initial linearization points of calibration parameters obviously deviated from the truth.

In contrast, less efforts were endeavored to the study of the observation update strategy to improve the filtering consistency, and for instance the open-source variants of MSCKF almost adopt the same update strategy proposed in [8]. Specifically, the filtering updates are performed when the features are not observed by the upcoming view any more, i.e., the tracked features along with the corresponding measurement constraints are not adopted until their disappearance. As compared in [7], the estimation accuracy of MSCKF cannot compete with the methods based on the sliding-window optimization, such as VINS-Mono [22], OKVIS [23], etc.

Within the MSCKF framework, [24] proposed a novel Kalman filter (PO-KF) by employing the pose-only representation [25][26]. It explicitly eliminates the 3D feature positions from the measurement equation. Consequently, the process of triangulating 3D feature points becomes unnecessary, enabling the immediate update of the system state using visual measurements from 3 camera frames. However, the paper implicitly posits that the immediate update strategy is only suitable for the PO-KF, which does not require triangulation for 3D point reconstruction, and lacks rigorous theoretical analyses of how immediate update contributes to enhancing the estimation performance. Partly inspired by the immediate measurement update strategy in

This paper was supported by in part by National Key R&D Program (2022YFB3903802), National Natural Science Foundation (62303310, 62273228, 62403315). (Qingchao Zhang and Wei Ouyang contributed equally. Corresponding author: Yuanxin Wu).
Authors' address: The authors are with Shanghai Key Lab of Navigation and Location Service, School of Electronic Information and Electrical Engineering, Shanghai Jiao Tong University, Shanghai 200240, China (email: yuanx_wu@hotmail.com).

PO-KF [24], this work investigated the immediate update strategy of MSCKF through the perspective of enlarging the accumulated information w.r.t. the observed features. Theoretical analyses reveal that the immediate update strategy can construct more observation constraints and performs more frequent state corrections, effectively depressing the linearization errors in contrast with the delayed update, which involves the early fixation of the linearization points. In addition, numerical simulations and experiments are performed to show the effectiveness of the proposed immediate update strategy in MSCKF. The main contributions of this work include:

(1) The immediate update strategy is proposed for MSCKF, and theoretical analyses from the perspective of information filtering indicate that immediate update leads to more accurate linearization points in essence.

(2) Simulations and experiments on open-source datasets show that the immediate update strategy improves the pose estimation accuracy about 28% in simulation, 29% in EuRoC dataset and 30% in KAIST VIO dataset over the MSCKF with delayed update.

(3) The proposed immediate update strategy for MSCKF can be readily combined with other modern filtering frameworks, such as the UKF, IEKF, and EqF, etc.

The remaining contents are organized as follows: Section II introduces the preliminaries about the MSCKF. Section III proposed the immediate update strategies and theoretical analyses are made from the perspective of information filtering. Section IV implements the numerical simulations to evaluate these update strategies, and experiments on open-source dataset are conducted in Section V. Section VI finally concludes this article.

## II. MSCKF AND ITS UPDATE STRATEGY

According to [8], [15], the preliminaries of the indirect Kalman filter and delayed update strategy used by MSCKF are provided here.

### A. Error-State Kalman Filter

The IMU state along with the mounting parameters defined in MSCKF is

$$\mathbf{x}_I = \left[ \mathbf{q}_G^{I\,T}, \left(\mathbf{v}_{GI}^G\right)^T, \left(\mathbf{p}_{GI}^G\right)^T, \mathbf{b}_g^T, \mathbf{b}_a^T, \mathbf{q}_C^{I\,T}, \left(\mathbf{p}_{IC}^I\right)^T \right]^T \quad (1)$$

in which the quaternion $\mathbf{q}_G^I$ denotes the frame rotation from the global reference frame (G) to the IMU body frame. $\mathbf{v}_{GI}^G, \mathbf{p}_{GI}^G \in \mathbb{R}^3$ denote the global-frame velocity and position vector of IMU w.r.t. the origin of $G$, respectively. $\mathbf{b}_g, \mathbf{b}_a \in \mathbb{R}^3$ are the gyroscope and accelerometer biases, respectively. The quaternion $\mathbf{q}_C^I$ denotes the frame rotation from the camera frame (C) to the IMU frame, and the mounting displacement $\mathbf{p}_{IC}^I$ denotes the relative position of camera w.r.t. the IMU. The initial mounting parameters between IMU and camera are usually calibrated by the Kalibr toolbox [27].

In this paper, $(\hat{\cdot})$ and $(\tilde{\cdot})$ denote the estimation and the measurement. The kinematic model of visual-inertial navigation system is given as

$$\begin{aligned}
\dot{\mathbf{q}}_G^I &= \frac{1}{2}\boldsymbol{\Omega}\left(\boldsymbol{\omega}_{GI}^I\right)\mathbf{q}_G^I, \\
\dot{\mathbf{v}}_{GI}^G &= \mathbf{C}_G^{I\,T}\mathbf{f}_{GI}^I + \mathbf{g}^G, \\
\dot{\mathbf{p}}_{GI}^G &= \mathbf{v}_{GI}^G, \\
\dot{\mathbf{b}}_g &= \mathbf{n}_{bg}, \dot{\mathbf{b}}_a = \mathbf{n}_{ba}, \\
\dot{\mathbf{q}}_C^I &= \mathbf{0}_{4\times 1}, \dot{\mathbf{p}}_{IC}^C = \mathbf{0}_{3\times 1}
\end{aligned} \quad (2)$$

where $\mathbf{C}_G^I$ is the rotation matrix from $G$ frame to $I$ frame, $\mathbf{n}_{bg}, \mathbf{n}_{ba}$ denote the noises of IMU bias. $\boldsymbol{\omega}_{GI}^I, \mathbf{f}_{GI}^I$ are related with the IMU measurements by $\tilde{\boldsymbol{\omega}}_{GI}^I = \boldsymbol{\omega}_{GI}^I + \mathbf{b}_g + \mathbf{n}_g$, $\tilde{\mathbf{f}}_{GI}^I = \mathbf{f}_{GI}^I + \mathbf{b}_a + \mathbf{n}_a$, in which, $\mathbf{n}_g, \mathbf{n}_a$ denote the noises of gyroscopes and accelerometers, respectively. The matrix $\boldsymbol{\Omega}$ is defined as

$$\boldsymbol{\Omega}\left(\boldsymbol{\omega}_{GI}^I\right) = \begin{bmatrix} -\boldsymbol{\omega}_{GI}^I \times & \boldsymbol{\omega}_{GI}^I \\ -\boldsymbol{\omega}_{GI}^{I\,T} & 0 \end{bmatrix}$$

where $(\cdot \times)$ denotes transforming a vector to a skew-symmetric matrix.

The error state of the indirect Kalman filter is defined as

$$\Delta \mathbf{x} = \left[\delta \boldsymbol{\theta}_I^T, \left(\delta \mathbf{v}_{GI}^G\right)^T, \left(\delta \mathbf{p}_{GI}^G\right)^T, \delta \mathbf{b}_g^T, \delta \mathbf{b}_a^T, \delta \boldsymbol{\theta}_C^T, \left(\delta \mathbf{p}_{IC}^I\right)^T \right]^T \quad (3)$$

in which, the Euler angle errors $\delta \boldsymbol{\theta}_I, \delta \boldsymbol{\theta}_C$ are defined by $\mathbf{q} \otimes \hat{\mathbf{q}} \approx \left[\delta \boldsymbol{\theta}^T/2, \ 1\right]^T$ and the other errors are defined by the true/measured state minus the estimate, such as $\delta \mathbf{p}_{GI}^G = \mathbf{p}_{GI}^G - \hat{\mathbf{p}}_{GI}^G$, and $\hat{\boldsymbol{\omega}}_{GI}^I = \tilde{\boldsymbol{\omega}}_{GI}^I - \hat{\mathbf{b}}_g$, $\hat{\mathbf{f}}_{GI}^I = \tilde{\mathbf{f}}_{GI}^I - \hat{\mathbf{f}}_{GI}^I$.

Then the linearized system model is formulated as

$$\Delta \dot{\mathbf{x}} = \mathbf{F}\Delta \mathbf{x} + \mathbf{G}\mathbf{w} \quad (4)$$

in which the Jacobian matrices and the continuous-time noises vector are given as

$$\mathbf{F} = \begin{bmatrix} -\hat{\boldsymbol{\omega}}_{GI}^I \times & 0 & 0 & -\mathbf{I} & 0 & 0 & 0 \\ 0 & -\left(\hat{\mathbf{C}}_G^I\right)^T \hat{\mathbf{f}}_{GI}^I \times & 0 & 0 & -\left(\hat{\mathbf{C}}_G^I\right)^T & 0 & 0 \\ 0 & 0 & \mathbf{I} & 0 & 0 & 0 & 0 \\ & & & \mathbf{0}_{15\times 21} & & & \end{bmatrix}$$

$$\mathbf{G} = \begin{bmatrix} -\mathbf{I} & 0 & 0 & 0 \\ 0 & -\left(\hat{\mathbf{C}}_G^I\right)^T & 0 & 0 \\ 0 & 0 & 0 & 0 \\ 0 & 0 & \mathbf{I} & 0 \\ 0 & 0 & 0 & \mathbf{I} \\ 0 & 0 & 0 & 0 \\ 0 & 0 & 0 & 0 \end{bmatrix}$$

$$\mathbf{w} = [\mathbf{n}_g^T, \mathbf{n}_a^T, \mathbf{n}_{bg}^T, \mathbf{n}_{ba}^T]^T$$

where $\mathbf{0}, \mathbf{I}$ denote the $3 \times 3$ zero matrix and identity matrix, respectively.

The states are propagated by the 4-th order Runge-Kutta method based on the differential equation (2), and the covariance propagation for the error-state in (3) is computed by

$$\bar{\mathbf{P}}_{II,k+1} = \mathbf{\Phi}_k \mathbf{P}_{II,k} \mathbf{\Phi}_k^T + \mathbf{Q}_k \quad (5)$$

where $(\bar{\cdot})$ represents the predicted state.

The state transition matrix and discrete-time noise matrix are given by [28]

$$\mathbf{\Phi}_k = \exp\left(\int_{t_k}^{t_{k+1}} \mathbf{F}(\tau) d\tau\right) \approx \mathbf{I} + \mathbf{F}_k \Delta t$$
$$\mathbf{Q}_k = \int_{t_k}^{t_{k+1}} \mathbf{\Phi}(t_k, \tau) \mathbf{G} \mathbf{Q} \mathbf{G} \mathbf{\Phi}(t_k, \tau)^\top d\tau$$
$$\approx \left(\mathbf{G}_k \mathbf{Q} \mathbf{G}_k + \mathbf{\Phi}_k \mathbf{G}_k \mathbf{Q} \mathbf{G}_k \mathbf{\Phi}_k^T\right) \Delta t / 2$$

where $\Delta t$ denotes the propagation time and the continuous-time process noise matrix $\mathbf{Q} = E[\mathbf{w}\mathbf{w}^T]$.

The pose of the camera can be obtained by the mounting relationship with the IMU as

$$\mathbf{C}_G^C = \mathbf{C}_I^C \mathbf{C}_G^I, \mathbf{p}_{GC}^G = \mathbf{p}_{GI}^G + \left(\mathbf{C}_G^I\right)^T \mathbf{p}_{IC}^I \quad (6)$$

Therefore, the Jacobian matrix between the camera pose error and the IMU state error is computed as

$$\mathbf{J}_{IC} = \begin{bmatrix} \hat{\mathbf{C}}_I^C & \mathbf{0} & \mathbf{0} & \mathbf{0} & \mathbf{0} & \mathbf{I} & \mathbf{0} \\ -\left(\hat{\mathbf{C}}_G^I\right)^T \mathbf{p}_{IC}^I \times & \mathbf{0} & \mathbf{I} & \mathbf{0} & \mathbf{0} & \mathbf{0} & \mathbf{0} \end{bmatrix} \quad (7)$$

If $N$ camera poses are enclosed in the MSCKF, the filter state will be augmented as

$$\mathbf{x} = \left[\mathbf{x}_I^T, \mathbf{q}_G^{C_1 T}, \left(\mathbf{p}_{GC_1}^G\right)^T, \cdots, \mathbf{q}_G^{C_N T}, \left(\mathbf{p}_{GC_N}^G\right)^T\right]^T \quad (8)$$

Hence, the augmented covariance matrix including the IMU states and camera poses can be initialized as

$$\mathbf{P} = \begin{pmatrix} \mathbf{P}_{II} & \mathbf{P}_{IC} \\ \mathbf{P}_{IC}^T & \mathbf{P}_{CC} \end{pmatrix} \quad (9)$$

in which, $\mathbf{P}_{CC} \in \mathbb{R}^{6N \times 6N}$ is the covariance for the errors of $N$ camera poses.

When a new camera pose is augmented, the covariance in (9) can be renewed by

$$\mathbf{P} = \begin{bmatrix} \mathbf{0}_{21+6N} \\ \mathbf{J}_{6 \times (21+6N)} \end{bmatrix} \mathbf{P} \begin{bmatrix} \mathbf{0}_{21+6N} \\ \mathbf{J}_{6 \times (21+6N)} \end{bmatrix}^T \quad (10)$$

where the coefficient matrix $\mathbf{J}_{6 \times (21+6N)} = \begin{bmatrix} \mathbf{J}_{IC} & \mathbf{0}_{6 \times 6N} \end{bmatrix}$.

*B. Delayed Update Strategy*

In MSCKF, a fixed number or window of camera poses is maintained and the features observed by the cameras in the window are applied to construct the observations. For the $j$-th feature measured by the $i$-th camera, the measurement model is given by projecting the feature observation to the image plane.

$$\mathbf{p}_{C_i f_j}^{C_i} = \begin{bmatrix} x_i^j & y_i^j & z_i^j \end{bmatrix}^T = \mathbf{C}_G^{C_i} \left(\mathbf{p}_{Gf_j}^G - \mathbf{p}_{GC_i}^G\right),$$
$$\mathbf{z}_i^j = \begin{bmatrix} x_i^j / z_i^j \\ y_i^j / z_i^j \end{bmatrix} = \mathbf{h}\left(\mathbf{p}_{Gf_j}^G, \mathbf{x}_{C_i}\right) \quad (11)$$

in which $\mathbf{h}\left(\mathbf{p}_{Gf_j}^G, \mathbf{x}_{C_i}\right) = \begin{bmatrix} \dfrac{e_1^T \mathbf{p}_{Gf_j}^G}{e_3^T \mathbf{p}_{Gf_j}^G} & \dfrac{e_2^T \mathbf{p}_{Gf_j}^G}{e_3^T \mathbf{p}_{Gf_j}^G} \end{bmatrix}^T$, $\mathbf{p}_{Gf_j}^G$ denotes the global 3D position of the $j$-th feature and $e_i$ denotes the $i$-th column of the $\mathbf{I}$.

This measurement model can be linearized w.r.t. the errors of camera pose and feature position as

$$\delta \mathbf{z}_i^j \approx \mathbf{H}_{C_i}^j \Delta \mathbf{x}_{C_i} + \mathbf{H}_{f_j}^i \Delta \mathbf{p}_j + \mathbf{n}_i^j \quad (12)$$

in which $\Delta \mathbf{x}_{C_i} = \left[\delta \boldsymbol{\theta}_{C_i}^T, \left(\delta \mathbf{p}_{GC_i}^G\right)^T\right]^T$, $\Delta \mathbf{p}_j = \mathbf{p}_{Gf_j}^G - \hat{\mathbf{p}}_{Gf_j}^G$, $\delta \mathbf{z}_i^j = \tilde{\mathbf{z}}_i - \mathbf{h}\left(\hat{\mathbf{p}}_{Gf_j}^G, \hat{\mathbf{x}}_{C_i}\right)$, $\mathbf{n}_i^j = \begin{bmatrix} \sigma_u & \sigma_v \end{bmatrix}^T$ and the corresponding Jacobian matrices can be computed as

$$\mathbf{H}_{C_i}^j = \begin{bmatrix} \cdots & \mathbf{J}_i^j \hat{\mathbf{p}}_{C_i f_j}^{C_i} \times & -\mathbf{J}_i^j \hat{\mathbf{C}}_G^{C_i} & \cdots \end{bmatrix}, \mathbf{H}_{f_j}^i = \mathbf{J}_i^j \hat{\mathbf{C}}_G^{C_i},$$
$$\mathbf{J}_i^j = \frac{1}{\left(z_i^j\right)^2} \begin{bmatrix} z_i^j & 0 & -x_i^j \\ 0 & z_i^j & -y_i^j \end{bmatrix} \quad (13)$$

Once the feature $f_j$ cannot be tracked by the upcoming view, the measurement matrices in (13) are to be computed using camera poses observing this feature. The stacked measurement residual is written by

$$\delta \mathbf{z}^j \approx \mathbf{H}_C^j \Delta \mathbf{x}_C + \mathbf{H}_{f_j} \Delta \mathbf{p}_j + \mathbf{n}^j \quad (14)$$

where the stacked measurement Jacobian matrices are

$$\mathbf{H}_C^j = \begin{bmatrix} \left(\mathbf{H}_{C_1}^j\right)^T, \cdots, \left(\mathbf{H}_{C_N}^j\right)^T \end{bmatrix}^T, \mathbf{H}_{f_j} = \begin{bmatrix} \left(\mathbf{H}_{f_j}^1\right)^T, \cdots, \left(\mathbf{H}_{f_j}^N\right)^T \end{bmatrix}^T \quad (15)$$

Since the position of feature point is not modelled in the filter's state (8), the left null space of $\mathbf{H}_{f_j}$, i.e., $\mathbf{A}^T \mathbf{H}_{f_j} = \mathbf{0}$, is commonly used to transform the residual in (14) as

$$\mathbf{A}^T \delta \mathbf{z}^j \approx \mathbf{A}^T \mathbf{H}_C^j \Delta \mathbf{x}_C + \mathbf{A}^T \mathbf{n}^j \quad (16)$$

Then, the resultant measurement innovations computed with all tracked feature points can be used in the routine Kalman filtering update. Since the measurement matrices and residuals are calculated on condition that the features are lost in the upcoming camera, this kind of update strategy is termed as the delayed update. In addition, updates can also be made when the window length of camera poses is larger than $N$. As conducted in S-MSCKF [15], the two camera poses on the window boundaries are selected to be marginalized, and the feature observations related with these camera poses are applied to the filtering update before removing them from the states.

III. ANALYSES ON IMMEDIATE UPDATE STRATEGY

The immediate update strategy of MSCKF is investigated in this section. Specifically, the positions of feature points are reconstructed online once a new camera pose and its feature observations are available. Then, the measurement constraints in (16) are formulated to perform the corrections on the IMU states and the camera poses within the window.

### A. Immediate Update Strategy

Comparing with the delayed update in MSCKF, the proposed immediate update strategy are exemplified in Fig. 1 using one tracked feature for simplicity. For this matched and tracked feature, immediate updates are sequentially performed from $t_3$ to $t_5$, while the delayed update is only triggered at $t_6$ until the feature is lost in the upcoming camera pose $C_6$. Specifically, the immediate updates sequentially construct 3 observation constraints w.r.t. $C_1C_2C_3$ at $t_3$, 4 observation constraints w.r.t. $C_1C_2C_3C_4$ at $t_4$, and 5 observation constraints w.r.t. $C_1C_2C_3C_4C_5$ at $t_5$. Nevertheless, the delayed update only constructs 5 observation constraints w.r.t. $C_1C_2C_3C_4C_5$ at $t_6$.

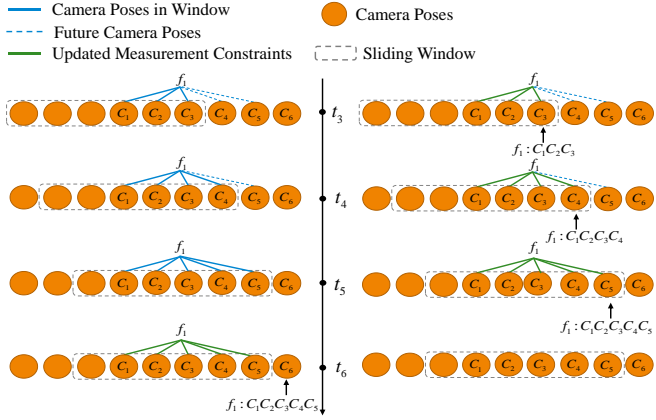

Fig. 1. The comparison of the delayed update (Left) and immediate update (Right).

Suppose the feature is lost in the $N+1$-th camera's observations at $t_{N+1}$, the filtering states for the delayed update is

$$\mathbf{x}_N = \left[ \mathbf{x}_I(t_N)^T, \mathbf{x}_{C_1}^T, \mathbf{x}_{C_2}^T, \cdots, \mathbf{x}_{C_N}^T \right]^T \quad (17)$$

in which $\mathbf{x}_{C_i}$ denotes the $i$-th camera poses, $\mathbf{x}_I(t_N)$ represents the IMU states at $t_N$.

According the information filtering theory, the updated covariance can be expressed as the inverse of information

$$\mathbf{P}_N = \left( \bar{\mathbf{P}}_N^{-1} + \mathbf{H}\mathbf{R}^{-1}\mathbf{H}^T \right)^{-1} \quad (18)$$

where $\bar{\mathbf{P}}_N$ is the propagated covariance matrix at $t_N$, and the covariance matrix of measurement noise is $\mathbf{R} = \mathbf{A}^T \mathrm{diag}\left( \left[\sigma_u^2, \sigma_v^2\right]_1, \cdots, \left[\sigma_u^2, \sigma_v^2\right]_N \right) \mathbf{A}$ and the Jacobian matrix $\mathbf{H} = \mathbf{A}^T \mathbf{H}_C^j$ has been given in (16). $\mathbf{H}\mathbf{R}^{-1}\mathbf{H}^T$ is also termed as the Fisher information from measurements[29].

The state correction can be computed by

$$\Delta \mathbf{x} = \left( \bar{\mathbf{P}}_N^{-1} + \mathbf{H}\mathbf{R}^{-1}\mathbf{H}^T \right)^{-1} \mathbf{H}^T \mathbf{R}^{-1} \mathbf{A}^T \mathbf{V}_N \quad (19)$$

in which, the stacked measurement residuals $\mathbf{V}_N = \left[ \mathbf{v}_1^T, \cdots, \mathbf{v}_N^T \right]^T$, $\mathbf{v}_i = \tilde{\mathbf{z}}_i - \mathbf{h}\left( \hat{\mathbf{p}}_{Gf_j}^G, \hat{\mathbf{x}}_{C_i} \right)$.

Note that the predicted IMU states and camera poses in (17) are propagated from the IMU states at $t_1$ to $t_i$ (i<=N) by

$$\mathbf{x}_I(t_i) = f_{t_i|t_1}\left( \mathbf{x}_I(t_1) \right), \mathbf{x}_{C_i} = g\left( \mathbf{x}_I(t_i), \mathbf{C}_I^C, \mathbf{p}_{IC}^C \right) \quad (20)$$

in which the function $f_{t_i|t_1}(\cdot)$ denotes the state propagation and $g(\cdot)$ denotes the map from the IMU pose to camera poses at $t_i$ as shown in (6).

In contrast, the immediate update starts at $t_3$ with the state vector as

$$\mathbf{x}_3 = \left[ \mathbf{x}_I(t_3)^T, \mathbf{x}_{C_1}^T, \mathbf{x}_{C_2}^T, \mathbf{x}_{C_3}^T \right]^T \quad (21)$$

The positions of feature points are reconstructed by the observations from $C_1C_2C_3$ and the state is corrected by

$$\mathbf{x}_3 = \bar{\mathbf{x}}_3 + \left( \bar{\mathbf{P}}_3^{-1} + \mathbf{H}_{C(1:3)} \mathbf{R}^{-1} \mathbf{H}_{C(1:3)}^T \right)^{-1} \mathbf{H}_{C(1:3)}^T \mathbf{R}^{-1} \mathbf{V}_3 \quad (22)$$

The updated covariance is given as

$$\mathbf{P}_3 = \left( \bar{\mathbf{P}}_3^{-1} + \mathbf{H}_{C(1:3)} \mathbf{R}^{-1} \mathbf{H}_{C(1:3)}^T \right)^{-1} \quad (23)$$

in which, $\bar{\mathbf{P}}_3$ is the propagated covariance up to $t_3$, $\mathbf{H}_{C(1:3)}$ is the stacked Jacobian matrix w.r.t. the camera poses $C_1C_2C_3$.

For the next update at $t_4$, positions of feature points are reconstructed by the observations from $C_1C_2C_3C_4$ and the filter state and covariance are updated.

$$\mathbf{x}_4 = \left[ \mathbf{x}_I(t_4)^T, \mathbf{x}_{C_1}^T, \mathbf{x}_{C_2}^T, \mathbf{x}_{C_3}^T, \mathbf{x}_{C_4}^T \right]^T \quad (24)$$

$$\mathbf{x}_4 = \bar{\mathbf{x}}_4 + \left( \bar{\mathbf{P}}_4^{-1} + \mathbf{H}_{C(1:4)} \mathbf{R}^{-1} \mathbf{H}_{C(1:4)}^T \right)^{-1} \mathbf{H}_{C(1:4)}^T \mathbf{R}^{-1} \mathbf{V}_4 \quad (25)$$

$$\mathbf{P}_4 = \left( \bar{\mathbf{P}}_4^{-1} + \mathbf{H}_{C(1:4)} \mathbf{R}^{-1} \mathbf{H}_{C(1:4)}^T \right)^{-1} \quad (26)$$

Note that the measurement Jacobian matrix $\mathbf{H}_{C(1:4)}$ and residuals $\mathbf{V}_4$ are computed by $\bar{\mathbf{x}}_4$, which is predicted from the corrected state $\mathbf{X}_3$ in (22).

If the covariance propagation between two camera poses is omitted for simplicity, i.e., $\mathbf{P}_{N-1} = \bar{\mathbf{P}}_N$, the updated covariance at $t_N$ would be approximated by

$$\begin{aligned}
\mathbf{P}_N^{-1} &= \mathbf{P}_{N-1}^{-1} + \mathbf{H}_{C(1:N)} \mathbf{R}^{-1} \mathbf{H}_{C(1:N)}^T \\
&= \mathbf{P}_{N-2}^{-1} + \mathbf{H}_{C(1:N-1)} \mathbf{R}^{-1} \mathbf{H}_{C(1:N-1)}^T + \mathbf{H}_{C(1:N)} \mathbf{R}^{-1} \mathbf{H}_{C(1:N)}^T \\
&\vdots \\
&= \mathbf{P}_1^{-1} + \sum_{i=3}^{N} \mathbf{H}_{C(1:i)} \mathbf{R}^{-1} \mathbf{H}_{C(1:i)}^T
\end{aligned} \quad (27)$$

in which, the stacked measurement Jacobian matrices $\mathbf{H}_{C(1:i)}$ are computed based on the propagated state from $t_{i-1}$ instead of that from the initial state $t_1$ in (17).

*B. Analyses*

Comparing (18) with (27), the immediate update strategy accumulates the measurement information from $N(N+1)/2-3$ observations up to $t_N$, whereas the delayed update only incorporates $N$ observations. One may argue that observations have been repeatedly constructed between the feature and the camera pose in the immediate update. Nevertheless, the camera poses and feature positions in $\mathbf{h}\left(\mathbf{p}_{Gf_j}^G, \mathbf{x}_{C_i}\right)$ are already corrected and reconstructed at each update time, making the measurement constraints different from the last update.

Remark 1. *The immediate update strategy constructs more measurement constraints than the delayed update, accumulating more information from feature observations and leading to smaller covariance matrix as given in* (27).

Besides, the state corrections are performed more frequently in the immediate update. As shown in the example, for the delayed update only one state update is implemented at $t_N$, but $N-2$ state corrections are made in the immediate update. It can be found that the IMU state for the immediate update at $t_N$ has been corrected $N-2$ times, while the IMU state in the delayed update is only corrected once as given by (19). The updated times for IMU state and camera poses in two update strategies are compared in Table I.

TABLE I

CORRECTED TIMES OF IMU STATE AND CAMERA POSES IN TWO UPDATE STRATEGIES

| Strategy | IMU | $C_{1:3}$ | $C_i (i>3)$ | $C_N$ |
|---|---|---|---|---|
| Delayed | 1 | 1 | 1 | 1 |
| Immediate | $N-2$ | $N-2$ | $N-i+1$ | 1 |

Remark 2. *In the immediate update strategy, the linearization points of the measurement Jacobian matrices and residuals become more accurate as more filtering updates are executed sequentially. In addition, the enhanced accuracy of IMU state and camera poses further contribute to the reconstruction of feature points and ameliorate the filtering consistency. Therefore, the accuracy of feature position* $\mathbf{p}_{Gf}^G$ *would tend to be better in immediate update, because more accurate linearization points and feature positions were conducive to improving the estimation performance.*

Note that more computations are required for the immediate update. It can be mitigated by constructing only part of the measurement constraints at each time. In this regards, the proposed strategy can be simplified to the 3-cam immediate update strategy as done similarly in [24], in which the feature observations relating with the first, middle and the last camera poses are exploited to build the measurement Jacobian matrices and residuals at each update time. Fig. 2 illustrates the first three steps of the 3-cam immediate update strategy. Besides, enclosing more than 3 feature observations at each update time is also feasible, such as 5-cam or 7-cam, and the strategy in (27) using all camera poses is named as all-cam immediate update hereafter.

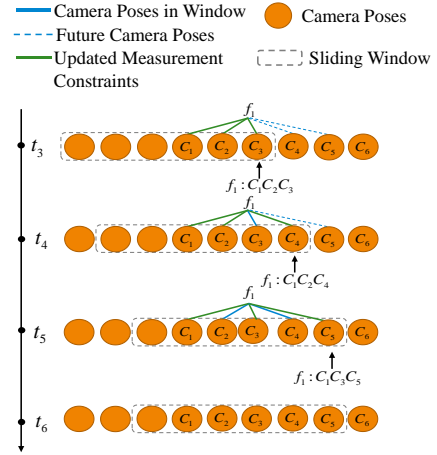

Fig. 2. The 3-cam immediate update strategy.

## IV. NUMERICAL SIMULATIONS

Numerical simulations in MATLAB are conducted to evaluate the effectiveness of the proposed immediate update strategy. We simulate an IMU whose body frame is defined as right-forward-up, and a camera is mounted pointing to the right direction of the IMU frame. The specifications of consumer-grade IMU are listed in Table II. The frequency of IMU is 100 Hz and the camera measurement is 10 Hz. As shown in Fig. 3, the feature points are randomly situated within a cylinder with 50m radius. The total number of feature points is 300 and the size of the camera image is 640×640 pixels. The intrinsic parameters of the camera are $f_x = 460$, $f_y = 460$, $c_x = 255$, $c_y = 255$. The standard deviation of the feature measurement error is 1 pixel. The relative position between the camera and IMU is $\mathbf{p}_{IC}^I = [5,4,3]^T$ cm.

Monte-Carlo simulations are conducted to compare three MSCKFs, i.e., the MSCKF with delayed update (MSCKF delayed) in [1][7][32], the MSCKF with all-cam immediate update (MSCKF all-cam) and the MSCKF with 3-cam immediate update (MSCKF 3-cam). The RMSEs of position and attitude errors across 50 runs are compared in Fig. 4 and the associated averaged RMSEs are listed in Table III. Results indicate that the all-cam and 3-cam MSCKF improves attitude estimation accuracy about 20%. Besides, the all-cam MSCKF outperforms the delayed MSCKF about 28% in the position accuracy, and the more efficient 3-cam MSCKF is 21% better in the position accuracy as well.

TABLE II

SPECIFICATIONS OF THE IMU IN SIMULATIONS

| Parameter | Gyroscope | Accelerometer |
|---|---|---|
| Constant bias | 50 deg/h | 100μg |
| Random walks | 0.6 deg/√h | 200μg/√Hz |

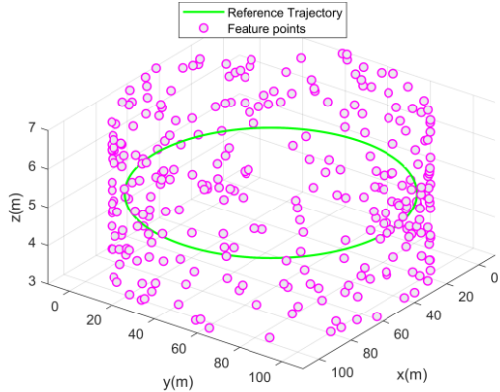

Fig. 3. The trajectory and feature points in simulation.

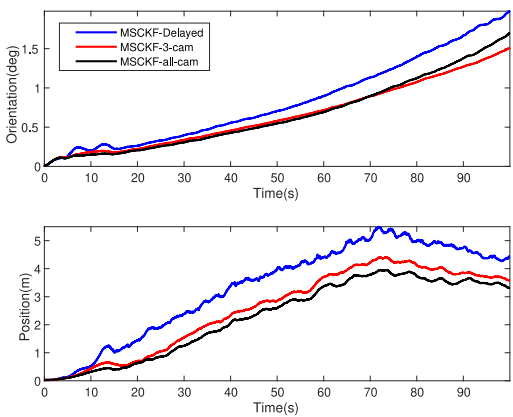

Fig. 4. The RMSEs of position and attitude errors in simulation.

TABLE III

THE AVERAGED RMSEs OF POSITION AND ATTITUDE ERRORS IN THE SIMULATIONS

| Methods | MSCKF Delayed | MSCKF 3-cam | MSCKF All-cam |
|---|---|---|---|
| Position (m) | 3.72 | 2.94 | 2.66 |
| Attitude (deg) | 0.99 | 0.77 | 0.80 |

## V. EXPERIMENTS

Two open-source datasets [30],[33] are applied to evaluate the proposed immediate update strategy. The algorithms are realized with C++ and the algorithms are run ten times as conducted in [19] to better evaluate the performance. The MSCKFs based on three update strategies (delayed, immediate all-cam, immediate 3-cam) use the same frontend and extrinsic/intrinsic parameters provided in the GitHub repository of [15], and the sliding window includes 20 camera poses. The averaged the lowest RMSEs of the absolute pose errors (APE) computed by the EVO tool [31] over ten runs are collaboratively used as the accuracy indexes.

### A. EuRoC Dataset

The EuRoC dataset is collected in indoor flight scenarios with three levels of difficulties. The frequencies of stereo images and IMU are 20 Hz and 200 Hz, respectively. Three MSCKFs are compared on EuRoC dataset and only the left images from the stereo camera are used. The settings of MSCKF are the same as [15] when evaluating the proposed immediate update strategy. The absolute pose errors are compared in Table IV, which shows that the all-cam/3-cam update strategies outperform the delayed strategy in 29% and 18% in MSCKF, respectively. The average time consumption in tackling each camera frame in EuRoC dataset is 0.0076s, 0.0128s, and 0.0216s for the delayed, 3-cam and all-cam update strategies, respectively.

Here, we also test the filtering performance of three algorithms after enlarging the number of observed features. In the front end of the MSCKF, the setting of grids and the min/max number of features in each grid are given as the default A{4,5,3,4} and the enlarged B{10,10,3,4}, respectively. In scenarios A and B, results of three algorithms are compared in Fig. 5. The all-cam and 3-cam based MSCKF still perform much better than the MSCKF with delayed update in scenario B. By increasing the number of observed features on this dataset, the estimation accuracy of three algorithms can be enhanced further and the all-cam MSCKF mostly performs the best.

TABLE IV

BEST/AVERAGED APE OF THREE ALGORITHMS OVER TEN RUNS ON EuRoC DATASET

| Dataset | MSCKF Delayed | MSCKF 3-cam | MSCKF All-cam |
|---|---|---|---|
| V1_01_easy | 0.106/0.137 | 0.060/0.083 | **0.057/0.076** |
| V1_02_medium | 0.138/0.180 | 0.085/0.107 | **0.059/0.094** |
| V1_03_difficult | 0.191/0.236 | 0.152/0.194 | **0.126/0.168** |
| V2_01_easy | 0.115/0.165 | **0.057/0.080** | 0.068/0.085 |
| V2_02_medium | 0.170/0.208 | 0.171/0.202 | **0.132/0.170** |
| MH_03_medium | 0.267/0.411 | 0.343/0.406 | **0.235/0.304** |
| MH_04_difficult | 0.374/0.479 | 0.316/0.425 | **0.304/0.389** |
| MH_05_difficult | 0.487/0.545 | 0.391/0.456 | **0.359/0.413** |
| Mean | 0.231/0.296 | 0.197/0.244 | **0.167/0.212** |

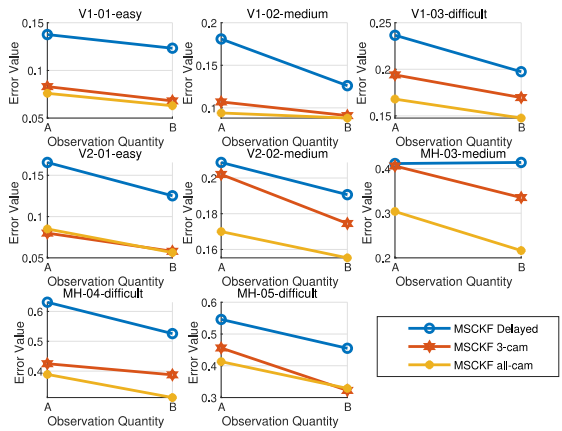

Fig. 5. The averaged RMSEs of APE over ten runs in each dataset with different numbers of features.

## B. KAIST VIO Dataset

The KAIST VIO dataset [33] was proposed for the UAVs platforms with NVIDIA Jetson processors, including four types of paths with different geometrical properties such as *circle*, *infinity*, *square* and *pure_rotation*. Each type of trajectory also contains three levels of difficulties, i.e., *normal* (normal speed with fixed heading), *fast* (high speed with fixed heading) and *head* (normal speed with rotational motion). The data were collected by Intel RealSense D435i camera in 30Hz and Pixhawk 4 mini IMU in 100Hz. The parameters of IMU noises are calibrated by the Allan variance, and the camera/IMU intrinsic and extrinsic parameters are calibrated by the Kalibr toolbox. The accurate ground truth for indoor trajectories is captured by an Opti-Track PrimeX 13 motion system. The compared three monocular VIO algorithms are all converged on the square_(normal/fast), infinity_(normal/head), and circle_(normal/fast/head) trajectories, and diverged for the challenging square_head, rotation data. The length of sliding window is 20 for KAIST VIO dataset, and the setting of grids and min/max number features in each grid are set as {4,5,3,4} as in open-source code of [33]. Table V lists the RMSEs of the evaluated monocular VIO algorithms. These results collectively indicate that the 3-cam and all-cam immediate update improve the estimation accuracy for 22% and 30%, respectively. The computational time for three algorithms is compared in Table VI, in which the 3-cam update can realize real-time processing but the all-cam update takes more time than the processing time 0.033s between consecutive camera frames. This fact indicates that a trade-off between computational efficiency and estimation accuracy should be reached by setting the constructed number of observations in the immediate update. In practice, the users are advised to try different number of the measurement constraints ranging from 3 to all camera poses and find a balance. And in our practice, 5-cam update is considered as a good choice, which both significantly enhances the filtering accuracy and meets the requirement of real-time process for the KAIST VIO dataset.

TABLE V

BEST/AVERAGED APE OF THREE ALGORITHMS OVER TEN RUNS ON KAIST VIO DATASET

| Dataset | MSCKF Delayed | MSCKF 3-cam | MSCKF All-cam |
|---|---|---|---|
| Squ_n | 0.252/0.308 | 0.162/0.212 | **0.148/0.170** |
| Squ_f | 0.125/0.149 | **0.083/0.113** | 0.088/0.113 |
| Inf_n | 0.280/0.328 | 0.223/0.290 | **0.175/0.245** |
| Inf_h | 0.397/0.410 | 0.3330/0.365 | **0.302/0.329** |
| Cir_n | 0.283/0.335 | 0.204/0.333 | **0.194/0.322** |
| Cir_f | 0.204/0.229 | 0.121/0.175 | **0.106/0.143** |
| Cir_h | 0.269/0.310 | 0.221/0.274 | **0.197/0.238** |
| Mean | 0.259/0.303 | 0.203/0.265 | **0.183/0.239** |

TABLE VI

THE PROCESSING TIME PER CAMERA FRAME OF THREE ALGORITHMS KAIST VIO DATASET

| Dataset | MSCKF Delayed | MSCKF 3-cam | MSCKF All-cam |
|---|---|---|---|
| Squ_n | 0.005 | 0.015 | 0.044 |
| Squ_f | 0.006 | 0.015 | 0.040 |
| Inf_n | 0.005 | 0.015 | 0.030 |
| Inf_h | 0.007 | 0.013 | 0.033 |
| Cir_n | 0.003 | 0.010 | 0.042 |
| Cir_f | 0.007 | 0.010 | 0.035 |
| Cir_h | 0.006 | 0.014 | 0.042 |
| Mean | 0.006 | 0.013 | 0.038 |

## VI. CONCLUSION

The delayed update has been applied in the MSCKF since its proposal. This work proposes an immediate update strategy through constructing the observation constraints upon the coming of feature measurements in each camera frame. Theoretical analyses of the immediate update against the delayed update are performed and reveal that the immediate update strategy helps construct more measurement constraints and accumulate more information from observations. Results of numerical simulations indicate that the immediate update strategy improves 28% of the pose estimation accuracy over the delayed update. Experiments on two datasets also show that the proposed strategy outperforms the MSCKF with delayed update by 29% on the EoRoC dataset and 30% on the KAIST VIO dataset. Admittedly, the improved estimation accuracy also brings about increased computation burden because of higher dimensional observations, and the trade-off of efficiency and accuracy should be well considered in practice.